\documentclass[10pt]{article} 
\usepackage[accepted]{tmlr}
\usepackage{graphicx}

\usepackage{hyperref}
\usepackage{amsmath,amsfonts,bm}
\usepackage{url}
\usepackage{cleveref}

\title{Protein Language Model Zero-Shot Fitness Predictions are Improved by Inference-only Dropout}

\author{\name Aditya Ravuri \email aditya.ravuri@cl.cam.ac.uk \\
      \addr
      University of Cambridge
      \AND
      \name Neil D. Lawrence \\
      \addr University of Cambridge
}

\begin{document}

\maketitle

\vspace{-1cm}
\begin{abstract}
Protein Language Models (PLMs) such as ESM2 \citep{esm2} have been shown to be capable of zero-shot prediction of critical scalar properties of proteins (``fitness'', \cite{zero-shot-plm}). In this work, we show that injecting a dropout layer \textbf{at inference time} between a PLM's featurizer/embedding layer and its transformer, and averaging its output akin to Monte-Carlo dropout \citep{mcd} increases zero-shot performance on a subset of the ProteinGym dataset \citep{proteingym}. This is the case even when the model was not trained with dropouts to begin with, and does not require retraining or finetuning of the PLM. A dropout of 0.1 seems performant across all models.
\end{abstract}

\section{Methodological Setup}

\begin{figure}[h]
    \centering
    \includegraphics[width=0.75\linewidth]{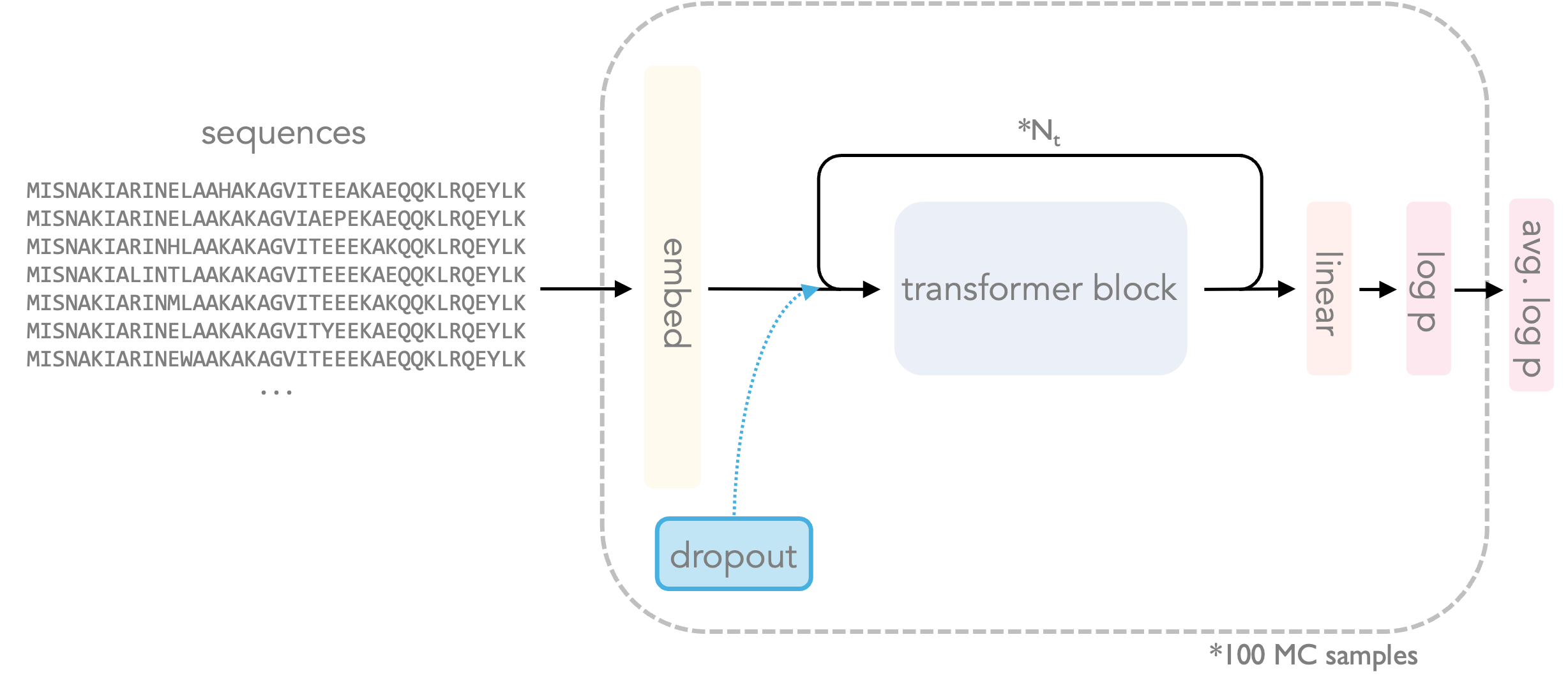}
    \caption{\small Graphical abstract: we simply introduce a dropout between the embedding layer and transformer block of a PLM, run many forward passes through the model and average the output log probabilities. These dropout-averaged outputs are more performant for zero-shot fitness prediction, even though the PLM was not trained with dropout.}
    \label{fig:graphical-abs}
\end{figure}

\textbf{The classical setup:} Let $\mathbf{s}$ denote a protein's amino acid sequence, represented as an integer sequence of length $n_a$. The problem concisely is, estimating a scalar property of the protein, represented by $y$. Outputs of PLMs can be used to construct a proxy $\hat{y}=\mathcal{T}(\mathbf{s})$, where $\mathcal{T} = \mathcal{S} \circ f$ is a composition of a scoring function $\mathcal{S}$ and a PLM $f$. The PLM $f$ maps the amino acid sequence $\mathbf{s} \in (\mathbb{N}^+)^{n_a}$ to a matrix of log-probabilities $\mathbf{L} = f(\mathbf{s}) \in \mathbb{R}^{n_a\times n_t}$, where $n_t$ is the number of possible tokens of the input\footnote{in practice, this is the number of proteinogenic amino acids (22) plus model-specific tokens representing a mask, etc.}. Given a family of mutations $\{\mathbf{s}_1, \mathbf{s}_2...\}$, we obtain $\{\mathbf{L}_1 = f(\mathbf{s}_1), \mathbf{L}_2 = f(\mathbf{s}_2), ...\}$, and use a scoring function $\mathcal S(\textbf{L}) = \sum_{ij} L_{ij}$ to construct scalar proxies for fitness $\{\hat{y}_1 = \mathcal S(\textbf{L}_1), \hat{y}_2 = \mathcal S(\textbf{L}_2), ...\}$\footnote{The $ij^{\text{th}}$ element of the matrix $\mathbf{L}$ represents log-probabilities corresponding to the values $\mathbf{s}_i$ can take, when these positions are masked, and a more appropriate scoring function should be constructed by quantifying how the joint log-probability of a sequence changes w.r.t. the wildtype sequence under the presence of a mutation. This is approximated in other work, e.g. by \cite{zero-shot-plm-nature}. We use a simplistic scoring function in this work due to its ease of implementation, and as we believe that this is also a valid proxy for out-of-domainness, as it's a proxy for model entropy given its input - discussed in \cref{sec:results}.}.

\textbf{Our method:} Our method for obtaining a proxy $\hat{y}$ follows the classical case, with one crucial difference, \textbf{we insert a dropout layer within the PLM}, between the embedding and transformer block, and obtain our proxy by averaging the outputs of the model, $\hat{y} = \mathcal S(\mathbb E_{d}(f_d(\mathbf{s})))$, where $d$ corresponds to the dropout. Our methods are illustrated graphically in \cref{fig:graphical-abs}.

For our experiments, we use the ESM2 suite of pretrained PLMs \citep{esm2}, and evaluate our proxies on a subset of the ProteinGym \citep{proteingym} DMS substitution dataset with 50 protein families. We use 100 Monte-Carlo samples for our dropout-averaging. For every protein family, we use the Spearman rank correlation coefficient (SRCC) to measure goodness-of-fit w.r.t. the true fitness values, and we report the median over all protein families within our data subset as the performance of a model.

\section{Results and Discussion}
\label{sec:results}

We believe that the scoring function we use is a proxy for entropy/inverse-confidence of the predicted distributions over tokens, and a measure of out-of-domainness of the input. Consider a categorical distribution with probabilities $\mathbf{p}$; as with the entropy, the quantity $\sum_i^K \log p_i$ is uniquely maximised when $\mathbf{p}$ is uniform. Given an out-of-domain example, we expect higher entropy in outputs, as observed in \cite{class-ood, zero-shot-mos}. Intuitively, given an OOD example, the entropy in model outputs would be higher, corresponding to a larger spread among its logits, leading to a higher mean of the logits. Out-of-domainness would correlate with fitness, if for example, fitness were to correspond to thermostability and most of the training data contained sequences from stable proteins.

\begin{figure}
    \centering
    \includegraphics[width=0.75\linewidth]{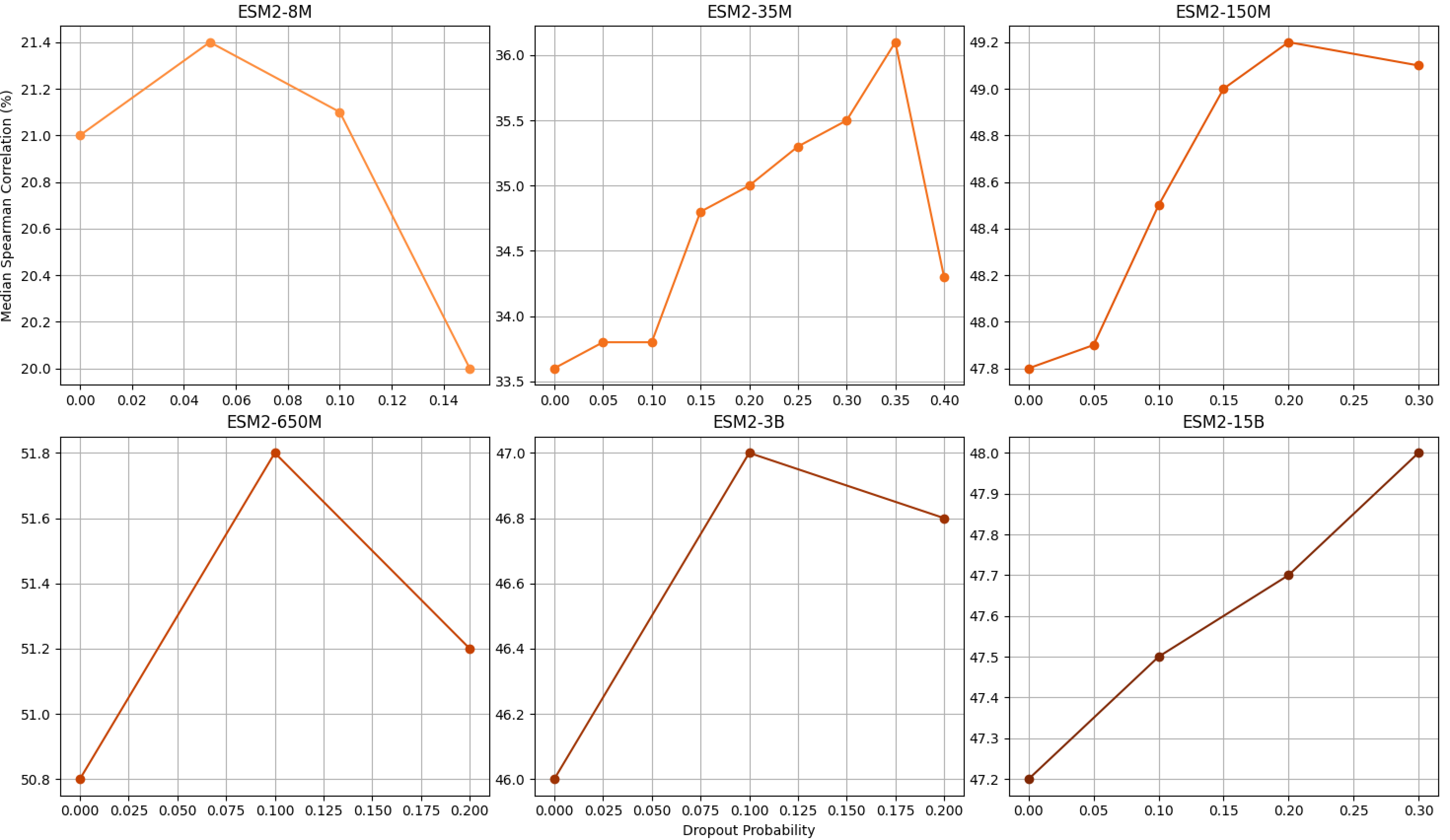}
    \caption{Graphs showing the zero-shot fitness performance of ESM2 with dropout added at inference-time only. We see an improvement in performance for every model size, but most strikingly in the case of the 35M parameter model. In the 150M and 15B cases, dropout layers had to be added to some of the early transformer layers too, for performance (the first fifth and first third respectively).}
    \label{fig:results}
\end{figure}

Our method is akin to Monte-Carlo dropout \citep{mcd}, although the model considered in this paper does not use dropout layers when training. Nevertheless, our results (illustrated in \cref{fig:results}) demonstrate that dropout-averaging improves zeroshot fitness prediction capabilities of PLMs (with dropout $=0$ corresponding to the untampered PLM's performance). We hypothesize that the increase in performance is due to this form of dropout-averaging increasing model calibration. A dropout of 0.1 seemes performant across all cases. Future work can expand our experimental suite to the full ProteinGym dataset, increase the number of gridpoints used for larger models (as for this work, we were limited by computational resources), and consider more appropriate scoring functions. Minimal reproducing \href{https://gist.github.com/InfProbSciX/f035ce643fba5fb304f036c1aed497fa}{code can be found on github.}

\newpage

\subsubsection*{Acknowledgments}
We'd like to thank Simon Mathis for a deep introduction to fitness prediction through their project on thermostability prediction, and Ferenc Huszár for insightful comments about transformers and for providing access to computational resources. AR would like to thank the Accelerate Programme for Scientific Discovery for supporting their PhD.

\bibliography{main}

\begin{thebibliography}{7}
\providecommand{\natexlab}[1]{#1}
\providecommand{\url}[1]{\texttt{#1}}
\expandafter\ifx\csname urlstyle\endcsname\relax
  \providecommand{\doi}[1]{doi: #1}\else
  \providecommand{\doi}{doi: \begingroup \urlstyle{rm}\Url}\fi

\bibitem[Gal \& Ghahramani(2016)Gal and Ghahramani]{mcd}
Yarin Gal and Zoubin Ghahramani.
\newblock Dropout as a bayesian approximation: Representing model uncertainty in deep learning, 2016.
\newblock URL \url{https://arxiv.org/abs/1506.02142}.

\bibitem[Hendrycks \& Gimpel(2018)Hendrycks and Gimpel]{class-ood}
Dan Hendrycks and Kevin Gimpel.
\newblock A baseline for detecting misclassified and out-of-distribution examples in neural networks, 2018.
\newblock URL \url{https://arxiv.org/abs/1610.02136}.

\bibitem[Hsu et~al.(2022)Hsu, Nisonoff, Fannjiang, and Listgarten]{zero-shot-plm-nature}
Chloe Hsu, Hunter Nisonoff, Clara Fannjiang, and Jennifer Listgarten.
\newblock Learning protein fitness models from evolutionary and assay-labeled data.
\newblock \emph{Nature Biotechnology}, 40\penalty0 (7):\penalty0 1114--1122, Jul 2022.
\newblock ISSN 1546-1696.
\newblock \doi{10.1038/s41587-021-01146-5}.
\newblock URL \url{https://doi.org/10.1038/s41587-021-01146-5}.

\bibitem[Lin et~al.(2023)Lin, Akin, Rao, Hie, Zhu, Lu, Smetanin, Verkuil, Kabeli, Shmueli, dos Santos~Costa, Fazel-Zarandi, Sercu, Candido, and Rives]{esm2}
Zeming Lin, Halil Akin, Roshan Rao, Brian Hie, Zhongkai Zhu, Wenting Lu, Nikita Smetanin, Robert Verkuil, Ori Kabeli, Yaniv Shmueli, Allan dos Santos~Costa, Maryam Fazel-Zarandi, Tom Sercu, Salvatore Candido, and Alexander Rives.
\newblock Evolutionary-scale prediction of atomic-level protein structure with a language model.
\newblock \emph{Science}, 379\penalty0 (6637):\penalty0 1123--1130, 2023.
\newblock \doi{10.1126/science.ade2574}.
\newblock URL \url{https://www.science.org/doi/abs/10.1126/science.ade2574}.

\bibitem[Meier et~al.(2021)Meier, Rao, Verkuil, Liu, Sercu, and Rives]{zero-shot-plm}
Joshua Meier, Roshan Rao, Robert Verkuil, Jason Liu, Tom Sercu, and Alex Rives.
\newblock Language models enable zero-shot prediction of the effects of mutations on protein function.
\newblock In M.~Ranzato, A.~Beygelzimer, Y.~Dauphin, P.S. Liang, and J.~Wortman Vaughan (eds.), \emph{Advances in Neural Information Processing Systems}, volume~34, pp.\  29287--29303. Curran Associates, Inc., 2021.
\newblock URL \url{https://proceedings.neurips.cc/paper_files/paper/2021/file/f51338d736f95dd42427296047067694-Paper.pdf}.

\bibitem[Notin et~al.(2023)Notin, Kollasch, Ritter, van Niekerk, Paul, Spinner, Rollins, Shaw, Orenbuch, Weitzman, Frazer, Dias, Franceschi, Gal, and Marks]{proteingym}
Pascal Notin, Aaron Kollasch, Daniel Ritter, Lood van Niekerk, Steffanie Paul, Han Spinner, Nathan Rollins, Ada Shaw, Rose Orenbuch, Ruben Weitzman, Jonathan Frazer, Mafalda Dias, Dinko Franceschi, Yarin Gal, and Debora Marks.
\newblock Proteingym: Large-scale benchmarks for protein fitness prediction and design.
\newblock In A.~Oh, T.~Naumann, A.~Globerson, K.~Saenko, M.~Hardt, and S.~Levine (eds.), \emph{Advances in Neural Information Processing Systems}, volume~36, pp.\  64331--64379. Curran Associates, Inc., 2023.
\newblock URL \url{https://proceedings.neurips.cc/paper_files/paper/2023/file/cac723e5ff29f65e3fcbb0739ae91bee-Paper-Datasets_and_Benchmarks.pdf}.

\bibitem[Ravuri et~al.(2024)Ravuri, Cooper, and Yamagishi]{zero-shot-mos}
Aditya Ravuri, Erica Cooper, and Junichi Yamagishi.
\newblock Uncertainty as a predictor: Leveraging self-supervised learning for zero-shot mos prediction.
\newblock In \emph{2024 IEEE International Conference on Acoustics, Speech, and Signal Processing Workshops (ICASSPW)}, pp.\  580--584, 2024.
\newblock \doi{10.1109/ICASSPW62465.2024.10626267}.

\end{thebibliography}
\bibliographystyle{tmlr}


\end{document}